\documentclass[sigconf]{acmart}
\renewcommand\footnotetextcopyrightpermission[1]{} 
\usepackage{bm}
\usepackage{booktabs}
\usepackage{amsmath}
\usepackage{makecell}
\usepackage{graphicx}
\usepackage{enumitem}
\usepackage{caption}
\usepackage{subfigure}
\usepackage{multicol}
\usepackage{multirow}
\usepackage{footnote}
\usepackage{balance}
\usepackage{tikz}
\makesavenoteenv{tabular}
\usepackage{latexsym}
\newcommand{\mat}[1]{{\bf #1}}   

\usepackage[T1]{fontenc}

\usepackage[utf8]{inputenc}
\AtBeginDocument{%
  \providecommand\BibTeX{{%
    \normalfont B\kern-0.5em{\scshape i\kern-0.25em b}\kern-0.8em\TeX}}}
\setcopyright{acmcopyright}
\copyrightyear{2018}
\acmYear{2018}
\acmDOI{XXXXXXX.XXXXXXX}

\acmConference[Conference acronym 'XX]{Make sure to enter the correct
  conference title from your rights confirmation emai}{June 03--05,
  2018}{Woodstock, NY}
\acmPrice{15.00}
\acmISBN{978-1-4503-XXXX-X/18/06}

\begin{document}

\title{Toward Understanding Bias Correlations for Mitigation in NLP}

\author{Lu Cheng}
\authornote{Both authors contributed equally to this research.}

\affiliation{%
  \institution{University of Illinois Chicago}
  \city{Chicago}
  \state{Illinois}
  \country{USA}
}
\email{lucheng@uic.edu}

\author{Suyu Ge}
\authornotemark[1]
\affiliation{%
  \institution{University of Illinois Urbana-Champaign}
  \city{Champaign}
  \state{Illinois}
  \country{USA}
}
\email{suyuge2@illinois.edu}
\author{Huan Liu}
\affiliation{%
  \institution{Arizona State University}
  \city{Tempe}
  \country{USA}}
\email{huanliu@asu.edu}

\renewcommand{\shortauthors}{Cheng, et al.}
\begin{abstract}
Natural Language Processing (NLP) models have been found discriminative against groups of different social identities such as gender and race. With the negative consequences of these undesired biases, researchers have responded with unprecedented effort and proposed promising approaches for bias mitigation. In spite of considerable practical importance, current algorithmic fairness literature lacks an in-depth understanding of the relations between different forms of biases. Social bias is complex by nature. Numerous studies in social psychology identify the ``generalized prejudice'', i.e., generalized devaluing sentiments across different groups. For example, people who devalue ethnic minorities are also likely to devalue women and gays. Therefore, this work aims to provide a first systematic study toward understanding bias correlations in mitigation. In particular, we examine bias mitigation in two common NLP tasks -- toxicity detection and word embeddings -- on three social identities, i.e., race, gender, and religion. Our findings suggest that biases are correlated and present scenarios in which independent debiasing approaches dominant in current literature may be insufficient. We further investigate whether jointly mitigating correlated biases is more desired than independent and individual debiasing. Lastly, we shed light on the inherent issue of debiasing-accuracy trade-off in bias mitigation. This study serves to motivate future research on joint bias mitigation that accounts for correlated biases.
\end{abstract}


\maketitle

\section{Introduction}
The increasing reliance on automated systems, e.g., systems helping decide who is hired, has led many people, especially the minorities and disadvantaged groups, to be unfairly treated. Take the two common tasks in Natural Language Processing (NLP) as examples: Toxicity classifiers are found to use demographic terms such as ``black'' as the key features and identify non-toxic tweets written in African American English as ``toxic'' with high confidence \cite{zhang2020demographics,zhou2021challenges}; word embeddings trained on human-generated corpus such as Google News and 100 years of text data with the US Census present strong stereotypes against women and ethnic minorities \cite{bolukbasi2016man,garg2018word}. 

\begin{figure}
\centering
  \resizebox{0.47\textwidth}{!}{\includegraphics{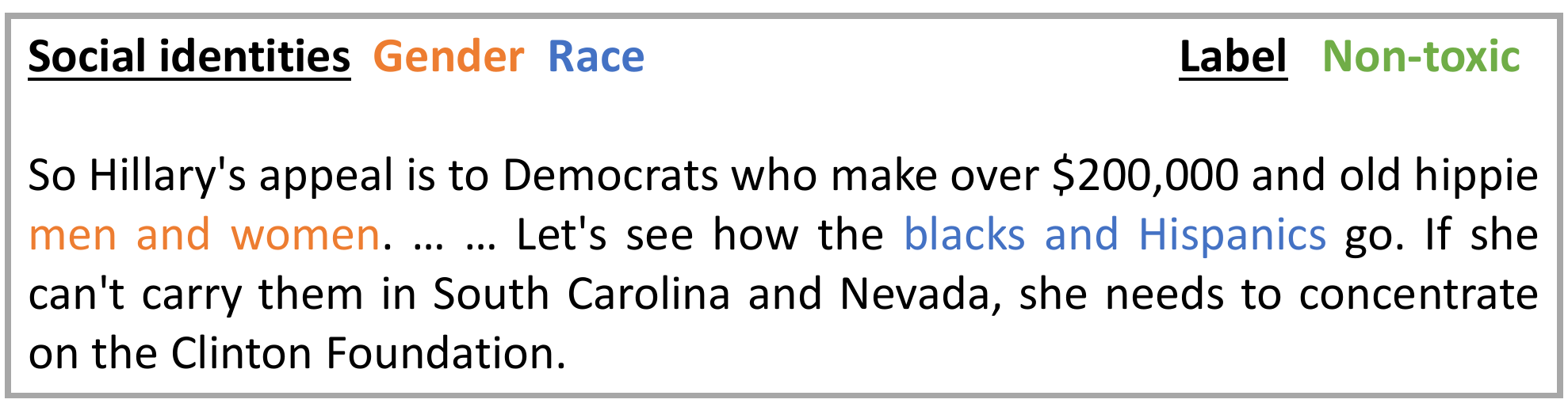}}
  \caption{A sample with annotated social identities from the benchmark dataset for toxicity detection, Jiasaw\protect\footnotemark.}
 \label{examp}
\end{figure}

These findings raised public awareness of the potential negative consequences of these powerful NLP systems. Tremendous efforts of industry, academia, and many other quarters have been put into understanding, detecting, and mitigating various social biases \cite{cheng2021socially}. In spite of considerable promising results, current algorithmic fairness literature might lack an in-depth understanding of the relations between different forms of biases. Typically, these social biases are treated independently in bias mitigation (e.g., mitigating gender bias is independent of mitigating racial bias). Social bias is complex by nature, in part due to its variety and the intersectionality (i.e., the complex, cumulative way in which the effects of multiple forms of biases combine, overlap, or intersect \cite{crenshaw2018demarginalizing}). As illustrated by the example of toxicity detection in Fig. \ref{examp}, different terms in a comment are labeled as being related to two social identities: gender and race. A biased toxicity classifier can incorrectly identify it as ``Toxic'' simply due to the presence of these identity terms, indicating gender and/or racial bias. Actually, findings about relations between biases have been documented numerous times across a range of cultural contexts \cite{akrami2011generalized,bierly1985prejudice}. Many scholars in social psychology have proposed that people who reject one social group tend to reject other groups \cite{bergh2016group,akrami2011generalized,bierly1985prejudice}. The latent factor behind different prejudices has been referred to the ``generalized prejudice'' \cite{allport1954nature}. With these issues considered, this work seeks to provide a systematic study towards understanding \textbf{bias correlations} in bias mitigation.
\footnotetext{https://www.kaggle.com/c/jigsaw-unintended-bias-in-toxicity-classification/data}

 \begin{figure}
\centering
  \resizebox{0.3\textwidth}{!}{\includegraphics{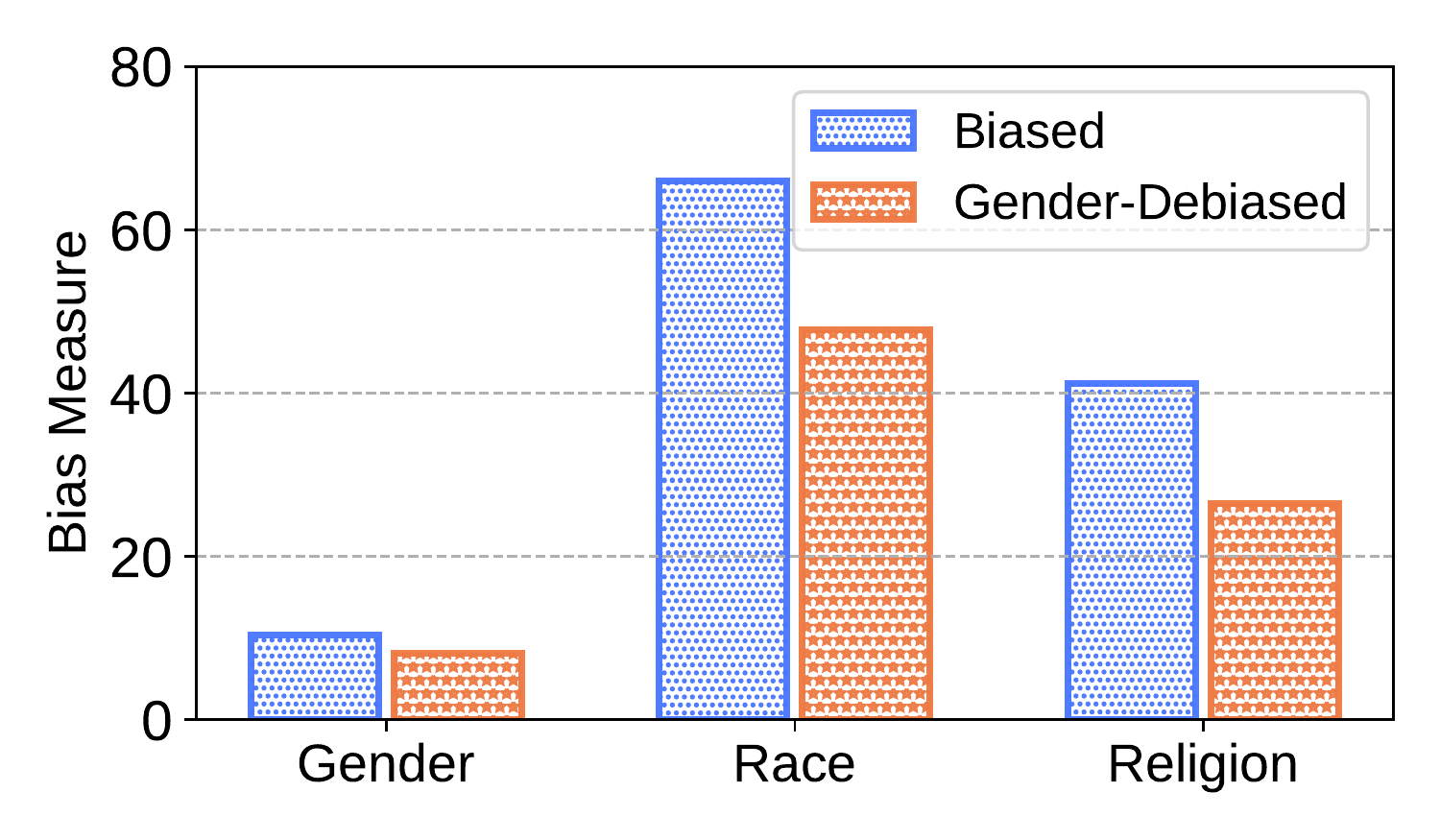}}
  \caption{An illustration of correlated biases in debiasing toxicity detection using the benchmark Jigsaw dataset. We observe that a gender-debiased toxicity classifier also reduces racial and religious biases.}
 \label{toxicity_intro}
\end{figure}

But first, why might it be \textit{acceptable} to consider independent debiasing? In a preliminary experiment on a benchmark dataset, we show that a gender-debiased toxicity classifier \cite{gencoglu2020cyberbullying} also reduces racial and religious biases in Fig. \ref{toxicity_intro}. This observation suggests a \textbf{positive} bias correlation (as also indicated by the previously mentioned sociological findings), therefore, an independent debiasing approach might still work to reduce the total bias. However, can there be \textbf{negative} bias correlations in bias mitigation? For example, when mitigating biases against various identities, a model debiased for one bias form may have negative influence on debiasing other forms of bias. When debiasing independently with negative bias correlations, failing to account for these correlations may render current bias mitigation approaches ineffective. Similar questions can be asked when debiasing at the data level. For example, \textit{will debiasing for gender negatively influence the results of debiasing for race and religion in word embeddings?} Therefore, in this work, we explore the possibility of negative bias correlations. Given its considerable practical importance, we further questions \textit{whether a joint debiasing approach can outperform conventional approaches for independent bias mitigation in terms of reducing the total bias?}

To provide an in-depth understanding toward bias correlations for bias mitigation in NLP, we examine multiple forms of biases at both \textit{data} and \textit{model} levels, on two representative NLP tasks -- word embeddings and toxicity detection. Particularly, we seek to answer the following research questions:
\begin{itemize}[leftmargin=*]
     \item \textbf{RQ. 1.} Are biases of different social identities (e.g., gender, race) correlated? How are they correlated across various tasks?
    \item \textbf{RQ. 2.} With potentially correlated biases, will a joint bias mitigation strategy outperform independent debiasing in terms of reducing the total bias of all social identities?
    \item \textbf{RQ. 3.} When jointly mitigating multiple biases, do we have to trade off accuracy/utility for debiasing? 
\end{itemize}
In \textbf{RQ. 3}, we shed light on a common issue of debiasing-accuracy trade-off in the bias and fairness literature for joint bias mitigation.
\section{Related Work}
\subsection{Bias Mitigation for Toxicity Detection}
\label{app:toxidebias}
The majority of existing works on debiasing toxicity detection addresses this problem via data augmentation~\cite{park2018reducing} and modified model training~\cite{zhou2021challenges,xia2020demoting}.
The pioneering work by \citet{dixon2018measuring} first proposed a data augmentation strategy to eliminate demographic biases. Specifically, it augmented the training dataset by adding external annotated data, which may cause semantic drift. Other data augmentation methods proposed to balance data distribution using existing datasets.
~\citet{park2018reducing} introduced gender swapping to equalize the number of male and female entities.
~\citet{mozafari2020hate} re-weighted training data to decrease the effects of n-grams that are highly correlated using class labels, and fine-tuned BERT using the re-weighted instances.
By attributing discrimination to selection bias, \citet{zhang2020demographics} proposed to mitigate biases in the training data by assuming a non-discrimination data distribution, and then reconstruct the distribution by instance weighting.

Another line of research mitigates biases during model training. For example,
\citet{vaidya2019empirical} trained a debiased toxicity classifier using adversarial learning.
They designed a multi-task learning model to jointly predict the toxicity label of a comment and the associated identities.
\citet{cheng2021mitigating} studied sequential bias mitigation for cyberbullying detection.
They proposed a reinforcement learning approach in which the toxicity classifier is an agent that interacts with the environment, comments sequence.
Also working on cyberbullying detection, \citet{gencoglu2020cyberbullying} modified the model training by imposing debiasing constraints on the classification objective. 
Both the debiasing approaches are model-agnostic, therefore, can be easily adapted to other model architectures.
In this work, we use the constraint-based approach \cite{gencoglu2020cyberbullying} as the base model.

Apart from data augmentation and modified model training, \citet{prost2019debiasing} investigated the efficacy of using debiased word embeddings for toxicity detection.
They found that the debiased embeddings actually amplified the amount of bias in toxicity detection and then proposed a modification to remove the gender information.
\citet{jin2020transferability} explored the transferability of debiased toxicity classifier.
They first fine-tuned an upstream model for bias mitigation and then fine-tuned it on downstream toxicity detection task.
A recent survey~\cite{zhou2021challenges} detected and individually mitigated three bias forms in toxicity detection: identity bias, swear word bias, and dialect bias. Another work provided a first systematic evidence on intersectional bias in datasets for hate speech and abusive language \cite{kim2020intersectional}. Though showing promising results, all previously mentioned research studied independent bias mitigation. 
\subsection{Word Embedding Bias}
In the development of NLP systems, different kinds of biases have been shown to exist in components such as training data, resources, and pre-trained models \cite{danks2017algorithmic,mehrabi2021survey}. Established research on word embedding bias has been focused on gender, e.g., \cite{zhao2019gender,basta2019evaluating,prost2019debiasing}. For example, the occupational stereotypes -- such as ``programmer'' is closer to ``man'' and ``homemaker'' is closer to ``woman'' -- were found in word2vec \cite{mikolov2013efficient} trained on the Google News dataset \cite{bolukbasi2016man}. In an extension to \cite{bolukbasi2016man}, \citet{manzini2019black} further identified racial and religious biases in word2vec trained on a Reddit corpus. Drawing from the Implicit Association Test (IAT) \cite{greenwald1998measuring} in social psychology, \citet{caliskan2017semantics} proposed the Word Embedding Association Test (WEAT) to examine the associations in word embeddings between concepts captured in IAT. WEAT aims to assess implicit stereotypes such as the association between female/male names and groups of words stereotypically assigned to females/males, e.g., arts vs. science. Some works (e.g., \cite{zhao2019gender,kurita2019measuring,basta2019evaluating}) also highlighted gender bias in various contextualized word embeddings such as BERT \cite{devlin2018bert}, ELMo \cite{Peters:2018}, and more recently, ALBERT \cite{lan2019albert}. Through five downstream applications related to emotion and sentiment intensity prediction, \citet{bhardwaj2021investigating} concluded that there is a ``significant dependence of the system's predictions on gender-particular words and phrases''. Bias in word embeddings can be further propagated to downstream tasks. For example, coreference systems were shown to link ``doctor'' to ``he'' and ``nurse'' to ``she'' \cite{zhao2018gender}; popular machine translation systems were found prone to gender biased translation errors \cite{stanovsky2019evaluating}. Human-like biases in BERT can also perpetuate gender bias in Gender Pronoun Resolution \cite{kurita2019quantifying}.

To mitigate gender bias, the post-processing methods (i.e., hard/soft debiasing) proposed in \cite{bolukbasi2016man} simply removed the component along the gender direction. For contextualized word embeddings, \citet{zhao2018gender} proposed to explicitly restrict gender information in certain dimensions when training a coreference system. Due to the demanding computational sources of this approach, an encoder-decoder model was proposed by \citet{kaneko2019gender} to re-embed existing pre-trained word embeddings. It still needs to train different encoder-decoders for different embeddings. However, these word embedding debiasing approaches have been scrutinized on several occasions \cite{gonen2019,blodgett2020language}. As shown with clustering in \cite{gonen2019}, debiased word embeddings still contain biases, therefore, existing bias removal techniques are insufficient to ensure a gender-neural modeling. Using such models to detect/mitigate biases will only project their remaining biases. Majority of existing works were also criticized not examining the impact of gender bias in real-world applications \cite{blodgett2020language}. In contrast to prior research in bias and fairness, this work aims to provide an in-depth understanding of how biases might interact with each other during bias removals. As human-like biases exist in the majority of word embeddings and debiasing approaches are unlikely to largely affect our findings, we use one of the pioneering works -- \cite{bolukbasi2016man} -- for the experimentation. Future research is warranted to investigate other debiasing approaches.   

In summary, current literature of both debiasing toxicity detection and word embeddings lacks an understanding of correlations among different bias forms. Therefore, this work complements prior research by providing the first systematic evidence on bias correlations in bias mitigation, including comprehensive quantitative and qualitative analyses as well as findings drawn from social psychology. We also show that biases can be negatively correlated, highlighting the need to develop joint bias mitigation strategies to simultaneously optimize for multiple biases to reduce the total bias.
\section{Preliminaries}
\subsection{Debiasing Toxicity Detection}
\label{sec:prelimi_toxicity}
A common bias mitigation strategy in toxicity detection adopts the metrics widely used to assess discrimination in classification tasks, i.e, False Negative Equality Difference (FNED) and False Positive Equality Difference (FPED)~\cite{dixon2018measuring}.
FNED/FPED is defined as the sum of deviations of group-specific False Negative Rates (FNRs)/False Positive Rates (FPRs) from the overall FNR/FPR.
Given $N$ demographic groups (e.g., female and male in gender), denote each group as $G_{i \in \{1,...,N\}}$, FNED and FPED are defined as:

\begin{small}
\begin{equation}
\begin{aligned}
    FNED&=\sum_{i \in \{1,...,N\}} |FNR-FNR_{G_i}|,\\
    FPED&=\sum_{i \in \{1,...,N\}} |FPR-FPR_{G_i}|.
\end{aligned}
\end{equation}
\end{small}
\noindent A debiased model is expected to have similar FNR and FPR for different groups belonging to the same identity, therefore, smaller FNED and FPED.
Ideally, the sum of FNED and FPED is close to zero for an unbiased model.

To reach this goal, a standard debiasing practice is the constrained model training~\cite{chen2020provably,zafar2017fairness}, which imposes debiasing constraints during the training process. It aims to reach equitable performances for different demographic groups of interest.
In this work, we employ the method in~\cite{gencoglu2020cyberbullying} as the base model. It simultaneously minimizes the deviation of each group-specific FNR/FPR from the overall FNR/FPR and the toxicity classification loss, i.e.,
\begin{small}
\begin{equation}
\begin{aligned}
\small
    &\min_{\theta}f_L(\theta)\\
    s.t.\quad \forall i, \text{ }
    |FN&R-FNR_{G_i}|<\tau_{FNR}\\
    |FP&R-FPR_{G_i}|<\tau_{FPR},\\
\end{aligned}
\end{equation}
\end{small}
\\
\noindent where $\tau_{FNR}$ and $\tau_{FPR}$ are the tolerances of group deviation (corresponding to biases) from overall FNRs and FPRs, respectively.
The model training process is then considered as a robust optimization problem. 
We use this approach as the base model since (1) it examines bias mitigation at the model level, therefore, complementing the task of debiasing word embeddings, which considers biases at the data level; and
(2) It is a generalizable and model-agnostic approach, and can be easily extended to the joint bias mitigation scenario, which we will detail in Sec. 4.1.

\subsection{Debiasing Word Embeddings}
A pioneering work in debiasing word embedding \cite{bolukbasi2016man} seeks to remove gender bias in word2vec \cite{mikolov2013efficient} by identifying the gender bias subspace. \citet{manzini2019black} further extended tit from a binary setting to the multi-class setting to account for other bias forms such as racial bias. For simplicity, this work uses the hard-debiasing approach \cite{bolukbasi2016man} designed for static word embeddings, future research is warranted to further examine approaches for debiasing contextualized word embeddings, such as \cite{zhao2019gender}. 

\noindent\textbf{Identifying the bias subspace.}
The bias subspace is identified by the \textit{defining sets} of words \cite{bolukbasi2016man} and words in each set represent different ends of the bias. The defining sets for gender can be \textit{\{she, he\}} and \textit{\{woman, man\}}. There are two steps to define a bias subspace: (1) computing the vector differences between the word embeddings of words in each set and the mean word embedding over the set; and (2) identifying the most $k$ significant components $\{\bm{b_1},\bm{b_2},...,\bm{b_k}\}$ of the resulting vectors using dimensionality reduction techniques such as Principal Component Analysis (PCA) \cite{abdi2010principal}.  

\noindent\textbf{Removing bias components.}
The next step completely or partially removes the subspace components from the embeddings, e.g., \textit{hard-debiasing} \cite{bolukbasi2016man}. 
For non-gendered words such as \textit{doctor} and \textit{nurse}, hard-debiasing method removes their bias components; for gendered words such as \textit{man} and \textit{woman}, it first centers their word embeddings and then equalizes the bias components. Formally, given a bias subspace $\mathcal{B}$ defined by a set of vectors $\{\bm{b_1},\bm{b_2},...,\bm{b_k}\}$, we get the bias component of an embedding in this subspace by 
\begin{equation}
\small
    \mat{w}_{\mathcal{B}}=\sum_{i=1}^k\big<\mat{w},\bm{b_i}\big>\bm{b_i}.
    \label{subspace}
\end{equation}
We then \textit{neutralize} word embeddings by removing the resulting component from non-gendered words: $\mat{w}'=\frac{\mat{w}-\mat{w}_{\mathcal{B}}}{\|\mat{w}-\mat{w}_{\mathcal{B}}\|},$
where $\mat{w}'$ are the debiased word embeddings. We further \textit{equalize} the gendered words in the equality set $E$. Specifically, we debias $\mat{w}\in E$ by 
\begin{equation}
\small
    \mat{w}'=(\bm{\mu}-\bm{\mu}_{\mathcal{B}})+\sqrt{1-\|\bm{\mu}-\bm{\mu}_{\mathcal{B}}\|^2}\frac{\mat{w}_{\mathcal{B}}-\bm{\mu}_{\mathcal{B}}}{\|\mat{w}_{\mathcal{B}}-\bm{\mu}_{\mathcal{B}}\|},
\end{equation}
where $\bm{\mu}=\frac{1}{|E|}\sum_{\mat{w}\in E}\mat{w}$ is the average embedding of the words in the set. $\bm{\mu}_{\mathcal{B}}$ denotes the bias component in the identified gender subspace and it can be obtained via Eq. \ref{subspace}. In real-world applications, the equality set is often the same set as the defining set \cite{manzini2019black}.
\section{Joint Bias Mitigation}
When the biases are correlated during debiasing (as we will show in the next section), an independent bias mitigation approach may be ineffective, or even amply other forms of bias, especially under negative bias correlations. With the variety of social biases, we essentially seek to reduce the total bias against all considered social identities. In this section, we introduce two simple joint bias mitigation strategies for debiasing toxicity detection and word embeddings, respectively.

\subsection{Joint Debiasing for Toxicity Detection}
\label{sec:approach_toxi}
The joint bias mitigation approach for toxicity detection is built upon the constraint-based fairness method described in Sec.~\ref{sec:prelimi_toxicity}. It can be easily adapted to other constraint-based fair machine learning models.
Let $T$=$\{gender, racial, religion\}$ be the set of considered social identities and $G_t$=$\{G_{t1}, G_{t2},..., G_{tj},... \}$ be the demographic group set with $t\in T$ (e.g., $G_{gender}$=$\{female, male\}$). 
A straightforward solution is to extend the debiasing method in Sec.~\ref{sec:prelimi_toxicity} by imposing uniform constraints on each demographic group $G_{tj}$.
Take identities \textit{gender} and \textit{race} as an example, we enforce the model to have similar FPR and FNR for \textit{male} and \textit{black}.
However, simply enforcing uniform performances across all identities may lead to sub-optimal solutions, due to the unique bias distributions and language characteristics within each social identity. 
For instance, it is more common to observe gender-related swear words in gender-biased comments than religion-biased comments.
A more desired solution needs to capture the unique characteristics of each bias form. Therefore, we propose to pair each demographic group exclusively with groups within the same identity when jointly debiasing for multiple biases.
That is, we enforce the model to have similar FPR and FNR for \textit{male} and \textit{female} in a binary gender, but not for \textit{male} and \textit{black} or \textit{male} and \textit{white}.
The \textit{joint debiasing constraint} is then defined as the following:
\begin{equation}
\small
\begin{aligned}
    &G_{tj} \in G_t \quad \forall t \in T\\
    |FN&R_{G_t}-FNR_{G_{tj}}|<\tau_{FNR},\\
    |FP&R_{G_t}-FPR_{G_{tj}}|<\tau_{FPR}. 
\end{aligned}
\end{equation}
Accordingly, the \textit{joint bias metric} is defined as the sum of $FNED_J$ and $FPED_J$ across all identities, where $FNED_J$ and $FPED_J$ are:
\begin{small}
\begin{equation}
\begin{aligned}
    FNED_J&=\sum_{t \in T}\sum_{G_{tj} \in G_t} |FNR_{G_t}-FNR_{G_{tj}}|,\\
    FPED_J&=\sum_{t \in T}\sum_{G_{tj} \in G_t} |FPR_{G_t}-FPR_{G_{tj}}|.\\
\end{aligned}
\end{equation}
\end{small}
\subsection{Joint Debiasing for Word Embeddings}
Given an identity $t\in T$, $n$ defining sets of word embeddings $\{D_{t1},D_{t2},...,D_{tn}\}$, and the word embedding $\mat{w}\in\mathbb{R}^d$ of word $w$, the bias subspace $\mathcal{B}_t$ is defined by the first $k$ components of the following PCA evaluation \cite{manzini2019black}:
\begin{equation}
\small
    \mathcal{B}_t=\textbf{PCA}\Big(\bigcup_{i=1}^{n}\bigcup_{\mat{w}\in D_{ti}}\mat{w}-\bm{\mu}_{ti}\Big),
\end{equation}
where $\bm{\mu}_{ti}=\frac{1}{|D_{ti}|}\sum_{\mat{w}\in D_{ti}}\mat{w}$ is a vector averaged over all word embeddings in set $i$. $\bigcup$ denotes concatenation by rows. 

Joint bias mitigation requires to simultaneously remove the bias components from a word embedding in all three identity subspaces. One solution is to take the mean of all bias subspaces as the joint bias subspace. However, the unique information of each bias form may be averaged out. Alternatively, we can concatenate the bias subspace $\mathcal{B}_t$ w.r.t. individual identities such that the resulting subspace $\mathcal{B}$ contains the significant components of all identity biases:
\begin{equation}
\small
    \mathcal{B}=\bigcup_{t\in T} \mathcal{B}_t.
\end{equation}
The joint bias subspace $\mathcal{B}\in \mathbb{R}^{3k\times d}$ allows the model to reserve the unique bias information of each identity in joint bias mitigation. 

\noindent\textbf{Quantifying Bias Removal.}
We use the mean average cosine distance (MAC) \cite{manzini2019black} to evaluate the individual bias in collections of words. Suppose we have a set of target word embeddings $\mathcal{S}$ that inherently contains certain form of social bias (e.g., \textit{Jew, Muslim}) and a set of attribute sets $\mathcal{A}=\{A_1, A_2,...,A_N\}$. $A_j$ consists of embeddings of words $\bm{a}$ that should not be associated with any word in $\mathcal{S}$ (e.g., \textit{violent, terrorist}). We define function $f(\cdot)$ to compute the mean cosine distance between $S_i\in \mathcal{S}$ and $\bm{a}\in A_j$:
\begin{equation}
\small
    f(S_i,A_j)=\frac{1}{|A_j|}\sum_{\bm{a}\in A_j}\cos(S_i,\bm{a}),
\end{equation}
where $\cos(S_i,\bm{a})=1-\frac{S_i\cdot\bm{a}}{\|S_i\|_2\cdot\|\bm{a}\|_2}$ measures the cosine distance, MAC is then defined as
\begin{equation}
\small
    \text{MAC}(\mathcal{S},\mathcal{A})=\frac{1}{|\mathcal{S}||\mathcal{A}|}\sum_{S_i\in \mathcal{S}}\sum_{A_j\in \mathcal{A}}f(S_i,A_j).
    \label{mac}
\end{equation}
A larger MAC score denotes greater bias removal.
\section{Experiments}
This section presents the major results of this work. In particular, we answer \textbf{RQ. 1 - RQ. 3} that seek to examine (1) the relations between different forms of bias; (2) the effectiveness of the proposed joint bias mitigation strategies across different NLP tasks; and (3) the debiasing-accuracy trade-off in joint bias mitigation.
\subsection{Task 1: Toxicity Detection}
We first describe the data source and basic experimental settings. We then discuss the main results.
\subsubsection{Data}
We use the Perspective API’s Jigsaw dataset of 403,957 samples with toxicity and identity annotations.
We use the same data split strategy as~\cite{gencoglu2020cyberbullying}, that is 70\% for training, 15\%
for validation, and 15\% for testing. The detailed statistics for each group are shown in Table~\ref{tab:toxicity_dataset}. All data in this study are publicly available and used under ethical considerations, see Appendix A.2. for the Ethics Statement.

\begin{table}[t]
	\centering
		\caption{Statistics of the Jigsaw dataset. We report the percentage of each social identity and the proportion of toxic samples within each group.
		}
	\resizebox{0.4\textwidth}{!}{
	\begin{tabular}{c|cccc}
    \Xhline{1pt}
    \textbf{Identity}&\multicolumn{2}{c|}{\textbf{Gender}}&\multicolumn{2}{c}{\textbf{Race}}\\
	Group&Male&\multicolumn{1}{c|}{Female}&Black&White\\
	\hline
	\% data&11.0\%&13.2\%&3.7\%&6.2\%\\
	\% toxicity&15.0\%&13.7\%&31.4\%&28.1\%\\
	\Xhline{1pt}
	\textbf{Identity}&\multicolumn{3}{c|}{\textbf{Religion}}&\multirow{2}{*}{\textbf{Overall}}\\
	Group&Christian&Jewish&\multicolumn{1}{c|}{Muslim}&\\
	\hline
	\% data&10.0\%&1.9\%&\multicolumn{1}{c|}{5.2\%}&38.1\%\\
	\% toxicity&9.1\%&16.2\%&\multicolumn{1}{c|}{22.8\%}&11.4\%\\
    \Xhline{1pt}
	\end{tabular}}
	\label{tab:toxicity_dataset}
\end{table}

\begin{table}[t]
	\centering
	\caption{Toxicity detection and bias mitigation performances of biased, individually debiased, and jointly debiased methods on the Jigsaw dataset.}
	\resizebox{\columnwidth}{!}{
	\begin{tabular}{c|ccc|cccc}
    \Xhline{1pt}
	\multirow{2}{*}{\textbf{Model}}& \multirow{2}{*}{\textbf{AUC}$\uparrow$}&\multirow{2}{*}{\textbf{F1}$\uparrow$}&\multirow{2}{*}{\textbf{Acc.}$\uparrow$}&\multicolumn{4}{c}{\textbf{Individual Bias Metric$\downarrow$}}\\
	 &&&&\textbf{Ge}& \textbf{Ra}& \textbf{Re}& \textbf{Total}\\
	\hline
	\textbf{Baseline}&\textbf{87.78}&51.21&85.07&10.36&65.98&41.16&117.50\\
	\hline
	\textbf{Ge}nder&87.73&54.93&89.83&\textbf{5.53}&47.79&26.54&\textbf{79.86}\\
	\textbf{Ra}ce&87.15&54.93&89.50&7.78&\textbf{39.58}&34.95&82.31\\
	\textbf{Re}ligion&87.71&55.42&89.36&11.26&66.53&\textbf{26.41}&104.20\\
	\Xhline{1pt}
	\multirow{2}{*}{\textbf{Model}}& \multirow{2}{*}{\textbf{AUC}$\uparrow$}&\multirow{2}{*}{\textbf{F1}$\uparrow$}&\multirow{2}{*}{\textbf{Acc.}$\uparrow$}&\multicolumn{4}{c}{\textbf{Joint Bias Metric$\downarrow$}}\\
	&&&&\textbf{Ge}& \textbf{Ra}& \textbf{Re}& \textbf{Total}\\
	\hline
	\textbf{Ge}+\textbf{Ra}&87.38&55.30&89.06&6.98&8.28&22.13&37.40\\
	\textbf{Ge}+\textbf{Re}&87.61&\textbf{55.43}&89.74&6.13&5.97&22.88&34.98\\
	\textbf{Ra}+\textbf{Re}&86.86&54.52&\textbf{90.10}&\textbf{4.44}&5.98&22.37&32.79\\
	\hline
	\textbf{Joint}&87.62&54.74&90.01&5.65&\textbf{4.72}&\textbf{20.29}&\textbf{30.65}\\
    \Xhline{1pt}
	\end{tabular}}
	\label{tab:toxicity_debias}
\end{table}

\begin{figure}[t]
\centering
  \resizebox{0.4\textwidth}{!}{\includegraphics{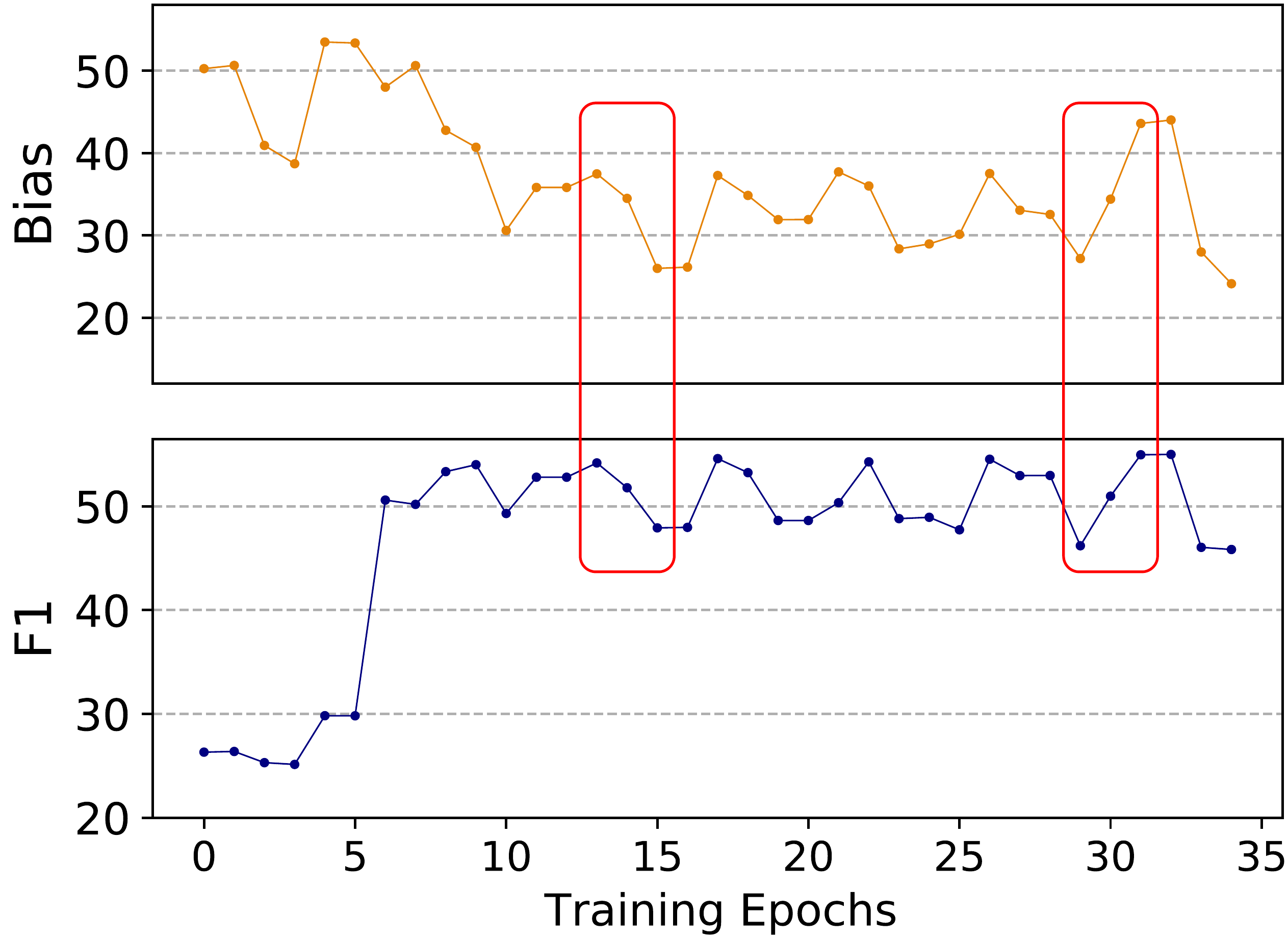}}
 \caption{Illustration of the inherent debiasing-accuracy trade-off in \textbf{Joint} during its training process.}
 \label{fig:trade-off}
\end{figure}

\subsubsection{Compared Approaches}

We compare the following models:
(1) biased \textbf{Baseline} model;
(2) models debiased for individual social identities (i.e., \textbf{Ge}nder, \textbf{Ra}ce and \textbf{Re}ligion);
(3) models debiased simultaneously for two identities (i.e., \textbf{Ge}+\textbf{Ra}, \textbf{Ge}+\textbf{Re}, and \textbf{Ra}+\textbf{Re});
(4) models debiased simultaneously for all three identities (i.e., \textbf{Joint}). 
Parameter settings can be found in Appendix A.1.

\subsubsection{Evaluation Metrics}
We use AUC, micro-F1, and accuracy (Acc.) scores as the evaluation metrics.
For bias mitigation, following \cite{dixon2018measuring,gencoglu2020cyberbullying}, we use the standard \textit{individual bias metric} introduced in Section~\ref{sec:prelimi_toxicity} for independent debiasing models (i.e., \textbf{Ge}nder, \textbf{Ra}ce, and \textbf{Re}ligion).
As the individual bias metric is not suitable when multiple forms of bias are present, we measure bias using the \textit{joint bias metric} described in Section~\ref{sec:approach_toxi} when debiasing for multiple forms of biases (i.e., \textbf{Ge}+\textbf{Ra}, \textbf{Ge}+\textbf{Re}, \textbf{Ra}+\textbf{Re}, and \textbf{Joint}).

\subsubsection{Results}
\label{sec:toxocity_result}
We have the following observations from Table~\ref{tab:toxicity_debias}: 

\noindent\textbf{RQ. 1. }Different forms of bias are inherently \textbf{\textit{correlated}} and the correlations tend to be \textbf{\textit{positive}}: mitigating one bias form will alleviate total bias w.r.t. all identities, e.g., comparing results of \textit{Baseline} with \textbf{Ge}nder, \textbf{Ra}ce, and \textbf{Re}ligion.
While aimed at reducing gender bias, \textbf{Ge}nder achieves competitive performance on mitigating religion bias compared to \textbf{Re}ligion.
Similar findings can be observed from models debiased simultaneously for two and three identities.
Therefore, with positive bias correlations in toxicity detection, an independent debiasing model can still help mitigate multiple forms of bias. This might explain the motivations to mitigate biases individually and independently. 

\noindent\textbf{RQ. 2. }
Although the biases are positively correlated, our proposed joint debiasing strategy still outperforms individual debiasing methods in terms of total bias mitigation.
As shown from the last two rows in Table~\ref{tab:toxicity_debias}, the \textbf{Joint} model effectively reduces multiple forms of bias simultaneously, achieving the least bias on race, religion, as well as total bias.
Instead of enforcing similar model performances among all groups, \textbf{Joint} only seeks for the closeness of FNR and FPR between groups within the same social identity, therefore better capturing the unique bias characteristics of individual identities.

\noindent\textbf{RQ. 3. }  
Another observation in Table \ref{tab:toxicity_debias} is that all the debiasing approaches (both individual and joint debiaisng) present competitive AUC and improved F1 and Accuracy scores. Similar observation can be found in previous works \cite{cheng2021mitigating,gencoglu2020cyberbullying}. Further, we show the total bias and F1 scores of \textit{Joint} over the training epochs in Fig.~\ref{fig:trade-off}. We can see from the highlighted parts that during training, the total bias changes in the same direction as changes of the F1 score, i.e., bias increases as F1 increases. This suggests that although bias mitigation approaches for toxicity detection (including the \textit{Joint}) can reduce bias and improve prediction performance, there is an inherent trade-off between debiasing and prediction performance during the model training. Therefore, debiasing the model alone may be insufficient, especially with multiple forms of bias. As shown in Table~\ref{tab:toxicity_dataset}, data distributions of different social identities are imbalanced, leading to more challenging optimization problem and further exacerbating the debiasing-accuracy trade-off. 
\subsection{Task 2: Word Embeddings}
Different from bias mitigation in toxicity detection that focuses on the design of NLP models, debiasing word embeddings typically is studied at the input data level. The purpose of this set of experiments is to answer \textbf{RQ. 1-3} from a data perspective. We first introduce the social bias, linguist data sources, and tasks considered for evaluation. We then present and discuss the major results. 
\begin{table*}
\begin{center}
\caption{MACs ($\uparrow$) w.r.t. different social identities for various versions of word embeddings. ``Target Identity'' denotes the identities of which we measure the MACs. ``Biased'' denotes the original biased word embeddings. ``H-Identity'' is the output after applying hard-debiasing to ``Biased'' to debias for a specific social identity. ``H-Identity-Identity'' is the result of sequential debiasing. Take the second row ``\textbf{Ra}ce'' as an example: Here, race is the target identity and gender is the debiased identity. We first measure MACs of race in Biased and H-Ge. After applying hard race debiasing to ``Biased'' and ``H-Ge'', we obtain ``H-Ra'' and ``H-Ge-Ra'' and then measure their race MACs (.925 and .892), respectively.
Under \textbf{Joint} is the race MAC after applying the proposed debiasing strategy to ``Biased'' to jointly reduce gender and racial biases.}
\label{rq1}
\begin{tabular}{c|cc|ccc||c|cc|ccc}
\Xhline{1pt}
Target Identity & Biased & H-Ge & H-Ra & H-Ge-Ra & Joint & Target Identity & Biased & H-Ge & H-Re   & H-Ge-Re & Joint \\ \hline
\textbf{Ra}ce            & .892   & .894 & \textbf{.925} & .892    & .924  & \textbf{Re}ligion        & .859   & .857 & .937 & .865 & \textbf{.941}  \\ \hline
Target Identity & Biased & H-Ra & H-Ge & H-Ra-Ge & Joint & Target Identity & Biased & H-Ra & H-Re   & H-Ra-Re & Joint \\ \hline
\textbf{Ge}nder           & .623   & .624 & \textbf{.695} & .654    & \textbf{.695}  & \textbf{Re}ligion        & .859   & .857 & .937   & .865    & \textbf{.940}  \\ \hline
Target Identity & Biased & H-Re & H-Ge & H-Re-Ge & Joint & Target Identity & Biased & H-Re & H-Ra   & H-Re-Ra & Joint \\ \hline
\textbf{Ge}nder           & .623   & .624 & .695 & .654    & \textbf{.790}  & \textbf{Ra}ce            & .892   & .890 & .925   & .889    & \textbf{.940}  \\ \Xhline{1pt}
\end{tabular}
\end{center}
\end{table*}
\begin{table}[ht!]
\begin{center}
\caption{MACs ($\uparrow$) w.r.t. different social identities after debiasing various word embeddings. ``Single'' indicates that we use hard debiasing to exclusively reduce bias of one social identity. Take the target identity (of which we present MACs) ``\textbf{Ge}nder'' as an example. ``H-Ge'' shows gender MACs after debiasing for gender on ``Biased''. Similarly, H-Re-Ge and Joint in the third-fourth columns are gender MACs after \textit{sequentially} and \textit{jointly} hard debiasing for religion and gender, respectively. H-Re-Ra-Ge and Joint in the fifth-sixth columns are gender MACs after \textit{sequentially} and \textit{jointly} hard debiasing for religion, race, and gender, respectively. * indicates statistically insignificant results.}
\label{rq2}
\begin{tabular}{c|c|cc|cc}
\Xhline{1pt}
\makecell{Target\\Identity}          & \multicolumn{1}{c|}{\textbf{Single}}    & \multicolumn{4}{c}{\textbf{Debiasing Multiple Social Identities}}                      \\ 
\Xhline{0.65pt}
\multirow{2}{*}{\textbf{Ge}nder}&H-Ge& H-Re-Ge & Joint &
                               H-Re-Ra-Ge     & Joint    \\ \cline{2-6} 
&              .695        & .654                 & .790           & .655               & \textbf{.794}     \\ 
\Xhline{0.65pt}
\multirow{2}{*}{\textbf{Ra}ce}& H-Ra  & H-Re-Ra      &Joint& H-Re-Ge-Ra&Joint    \\ \cline{2-6} 
&.925&   .889                 & \textbf{.940}  & .891               & .880*              \\ 
\Xhline{0.65pt}
\multirow{2}{*}{\textbf{Re}ligion}&H-Re& H-Ge-Re    & Joint &H-Ge-Ra-Re&Joint\\  \cline{2-6} 
&.937&   .865                 & \textbf{.941}  & .865               & .930              \\ \Xhline{1pt}
\end{tabular}
\end{center}
\end{table}
\subsubsection{Social Bias and Linguistic Data}
For the embedding language corpus, we use the L2-reddit corpus \cite{rabinovich2018native}, a collection of Reddit posts and comments by both native and non-native English speakers. In this work, we exclusively include data collected from the United States as it produces comparatively larger amount of studies about social biases. The initial biased word embeddings are obtained by training word2vec on approximately 56 million sentences. For lexicons for gender bias, we use vocabularies curated by \cite{bolukbasi2016man} and \cite{caliskan2017semantics}. For racial and religious biases, we use the list of lexicons curated by \cite{manzini2019black}. 
\subsubsection{Evaluation Tasks} 
We design and conduct the following experiments. For bias correlations (\textbf{RQ. 1}), we first quantify the bias removals (i.e., MACs) of the other two social identities after debiasing for one identity. For example, after we apply hard-debiasing to reducing gender bias, we measure MACs of race and religion, which are then compared with MACs of race and religion in original biased word embeddings. We denote race and religion as Target Identity and gender as Debiased Identity. As only embeddings of non-gendered words are debiased when the debiased identity is \textbf{Ge}, the changes of MACs w.r.t. race and religion are expected to be close to zero. As a result, we further examine the influence of debiasing for gender on the subsequent debiasings w.r.t. race and religion. For example, \textit{will mitigating racial bias for the gender-debiased word embeddings less effective than for the original biased embeddings? }

For \textbf{RQ. 2}, we compare \textit{Joint} with baselines that apply the hard-debiasing method to the original embeddings \textbf{sequentially}. For instance, to mitigate both gender and racial biases, the baseline debiases the gender-debiased word embeddings for race whereas \textit{Joint} simultaneously debiases the original word embeddings for race and gender. We also perform a paired $t$-test on the distribution of average cosine distance used to compute MAC. Unless otherwise noted, results of MAC below are statistically significant at level 0.05. For \textbf{RQ. 3}, we follow previous works (e.g., \cite{manzini2019black,bolukbasi2016man,zhao2019gender}) to evaluate the utility of debiased word embeddings on common downstream NLP tasks. We consider NER tagging, POS tagging, and POS chunking, as in \cite{manzini2019black}. Data are provided by the CoNLL 2003 shared tasks \cite{sang2003introduction}. There are two evaluation paradigms: replacing the biased embeddings with the debiased ones or retraining the model on debiased embeddings. A desired debiasing strategy should reduce the biases in word embeddings and remain their utilities.

In summary, we use the following notations to represent various versions of word embeddings. The original biased word embeddings are denoted as \textbf{Biased} and word embeddings after hard-debiasing are denoted as H-Identity. For example, the output of hard gender debiasing is denoted as \textbf{H-Ge}. Similarly, we can get \textbf{H-Ra} and \textbf{H-Re}. For output after sequentially applying hard-debiasing, we denote it as H-Identity-Identity. For example, the output of hard race debiasing on \textbf{H-Ge} is denoted as \textbf{H-Ge-Ra}. The joint bias mitigation approach is denoted as \textbf{Joint}.
\begin{table*}[t]
\setlength\tabcolsep{1pt}
\begin{center}
\caption{Top five analogies of \{\textit{man, woman}\} generated by various word embeddings after being \textbf{debiased for Ge}nder. Each entry below can be interpreted as ``man is to XX as woman is to XX''.}
\label{analogy}
\resizebox{\textwidth}{!}{
\begin{tabular}{c|c|c|c|c}
\Xhline{1pt}
\multirow{2}{*}{\textbf{H-Ge}} & \multicolumn{2}{c|}{\textbf{Re}ligion + \textbf{Ge}nder}                          & \multicolumn{2}{c}{\textbf{Re}ligion + \textbf{Ra}ce + \textbf{Ge}nder}                    \\ \cline{2-5} 
                                  & \textbf{H-Re-Ge}           & \textbf{Joint}           & \textbf{H-Re-Ra-Ge}                 & \textbf{Joint}         \\ \hline
(executive, executive)            & \textbf{(chairman, secretary)} & (homemaker, homemaker)   & (executive, executive)         & (stylist, stylist)     \\
(homemaker, homemaker)            & (executive, executive)         & (stylist, stylist)       & \textbf{(chairman, secretary)} & (homemaker, homemaker) \\
(manager, manager)                & (homemaker, homemaker)         & (manager, manager)       & (homemaker, homemaker)         & (clerk, clerk)         \\
(clerk, clerk)                    & (secretary, secretary)         & (programmer, programmer) & (secretary, secretary)         & (executive, executive) \\
(secretary, secretary)            & (manager, manager)             & (supervisor, supervisor) & (manager, manager)             & (singer, singer)       \\ \Xhline{1pt}
\end{tabular}}
\end{center}
\end{table*}
\begin{table*}
\caption{\textit{Embedding Matrix Replacement:} Utility of word embeddings debiased in NER Tagging, POS Tagging, and POS Chunking. \textbf{Seq} and \textbf{Joint} denote sequential and joint debiasing, respectively. $\Delta$ denotes the change of a metric before and after debiasing.}
\centering
\label{downstream1}
\resizebox{\textwidth}{!}{
\begin{tabular}{|c|c|c|c|c|c|c|c|c|c|}
\hline
\multirow{2}{*}{\makecell{\textbf{Target} \\\textbf{Identity}} } & \multicolumn{3}{c|}{\textbf{Re}ligion}   & \multicolumn{3}{c|}{\textbf{Re}ligion +\textbf{Ge}nder} & \multicolumn{3}{c|}{\textbf{Re}ligion + \textbf{Ge}nder + \textbf{Ra}ce} \\ \cline{2-10} 
                  & NER Tagging & POS Tagging & POS Chunking & NER Tagging   & POS Tagging    & POS Chunking   & NER Tagging      & POS Tagging      & POS Chunking     \\ \hline
Biased F1         & .9930       & .9677       & .9968        & .9930         & .9677          & .9968          & .9930            & .9677            & .9968            \\ \hline
                  & \multicolumn{3}{c|}{\textit{NA}}                 & \textbf{Seq / \textbf{Joint}}    & \textbf{Seq / \textbf{Joint}}    & \textbf{Seq / \textbf{Joint}}     & \textbf{Seq / \textbf{Joint}}       & \textbf{Seq / \textbf{Joint}}       & \textbf{Seq / \textbf{Joint}}      \\ \hline
$\Delta$ F1                & +.007        & -.026       & +.003         & +.004 / +.004    & -.011 / -.009   & +.004 / +.005     & +.004 / +.004       & -.012 / -.013     & +.004 / +.005       \\ \hline
$\Delta$ Precision         & .0          & -.029       & .0           & .0 / .0        & -.023 / -.019   & .0 / .0         & .0 / .0           & -.026 / -.026     & .0 / .0           \\ \hline
$\Delta$ Recall            & +.025        & -.073       & +.012         & +.015 / +.015    & -.020 / -.018   & +.016 / +.017     & +0.014 / +.016      & -.021 / -.025     & +.014 / +.019       \\ \hline
\end{tabular}}
\end{table*}
\begin{table*}
\centering
\caption{\textit{Model Retraining:} Utility of word embeddings debiased in NER Tagging, POS Tagging, and POS Chunking. }
\label{downstream2}
\resizebox{\textwidth}{!}{
\begin{tabular}{|c|c|c|c|c|c|c|c|c|c|}
\hline
\multirow{2}{*}{\makecell{\textbf{Target} \\\textbf{Identity}} } & \multicolumn{3}{c|}{\textbf{Re}ligion}   & \multicolumn{3}{c|}{\textbf{Re}ligion +\textbf{Ge}nder} & \multicolumn{3}{c|}{\textbf{Re}ligion + \textbf{Ge}nder + \textbf{Ra}ce} \\ \cline{2-10}
                  & NER Tagging & POS Tagging & POS Chunking & NER Tagging    & POS Tagging   & POS Chunking   & NER Tagging      & POS Tagging      & POS Chunking     \\ \hline
Biased F1         & .9930       & .9677       & .9968        & .9930          & .9677         & .9968          & .9930            & .9677            & .9968            \\ \hline
                  & \multicolumn{3}{c|}{NA}  
                   & \textbf{Seq / \textbf{Joint}}    & \textbf{Seq / \textbf{Joint}}    & \textbf{Seq / \textbf{Joint}}     & \textbf{Seq / \textbf{Joint}}       & \textbf{Seq / \textbf{Joint}}       & \textbf{Seq / \textbf{Joint}}      \\ \hline
$\Delta$ F1                & +.007        & -.011       & +.003         & +.004 / +.004     & +.003 / -.011   & +.004 / +.005     & +.004 / +.004       & -.010 / -.007     & +.004 / +.005       \\ \hline
$\Delta$ Precision         & .0          & -.011       & .0           & .0 / .0         & +.006 / -.023   & .0 / .0         & .0 / .0           & -.018 / -.007     & .0 / .0           \\ \hline
$\Delta$ Recall            & +.025        & -.031       & +.012         & +.015 / +.015     & +.005 / -.019   & +.016 / +.017     & +.014 / +.016      & -.021 / -.023     & +.014 / +.019       \\ \hline
\end{tabular}}
\end{table*}
\subsubsection{Results}
\textbf{RQ. 1.} We have several observations from the results in Table \ref{rq1}: (1) When debiasing for various identities sequentially, hard-debiasing tends to be less effective w.r.t. the second and third identities, or even amplify the biases against those identities. That is, sequential debiasing potentially has \textbf{\textit{negative}} influence on the effectiveness of the hard-debiasing approach. For example, in the second row (\textbf{Ra}ce) in Table \ref{rq1}, \textit{H-Ra} (.925) shows larger race MAC than \textit{H-Ge-Ra} (.892). This indicates that directly debiasing for race on \textit{Biased} removes more racial bias compared to debiasing for gender and race sequentially. (2) As expected, when we debias for one identity, the amount of bias of other identities barely changes, e.g., comparing the results in the second and the third columns in Table \ref{rq1}. For example, race MACs of \textit{Biased} (.892) and \textit{H-Ge} (.894) are close. This is because the original hard debiasing approach and the bias measure (MACs) rely on the pre-defined vocabulary sets that do not intersect. Future research is warranted to look into vocabularies associated with multiple biases. (3) To further examine the influence of sequential debiasing, we extend results in Table \ref{rq1} and debias for multiple social identities. Results in Table \ref{rq2} show supporting evidence for findings in (1). Take the target identity \textbf{Ge}nder as an example: in Row 3, both gender MACs for two sequential debiasings, i.e., H-Re-Ge (.654) and H-Re-Ta-Ge (.655), are smaller (therefore, less bias removal) than directly debiasing for gender (\textit{H-Ge}: .695). Note that although there are multiple potential debiasing sequences, we find consistent conclusions over sequential hard debiasing with all different orders.  

\begin{figure*}[t]
    \centering
\subfigure[Toxicity proportion in \textit{Jigsaw}.]{\resizebox{0.28\textwidth}{!}{\includegraphics{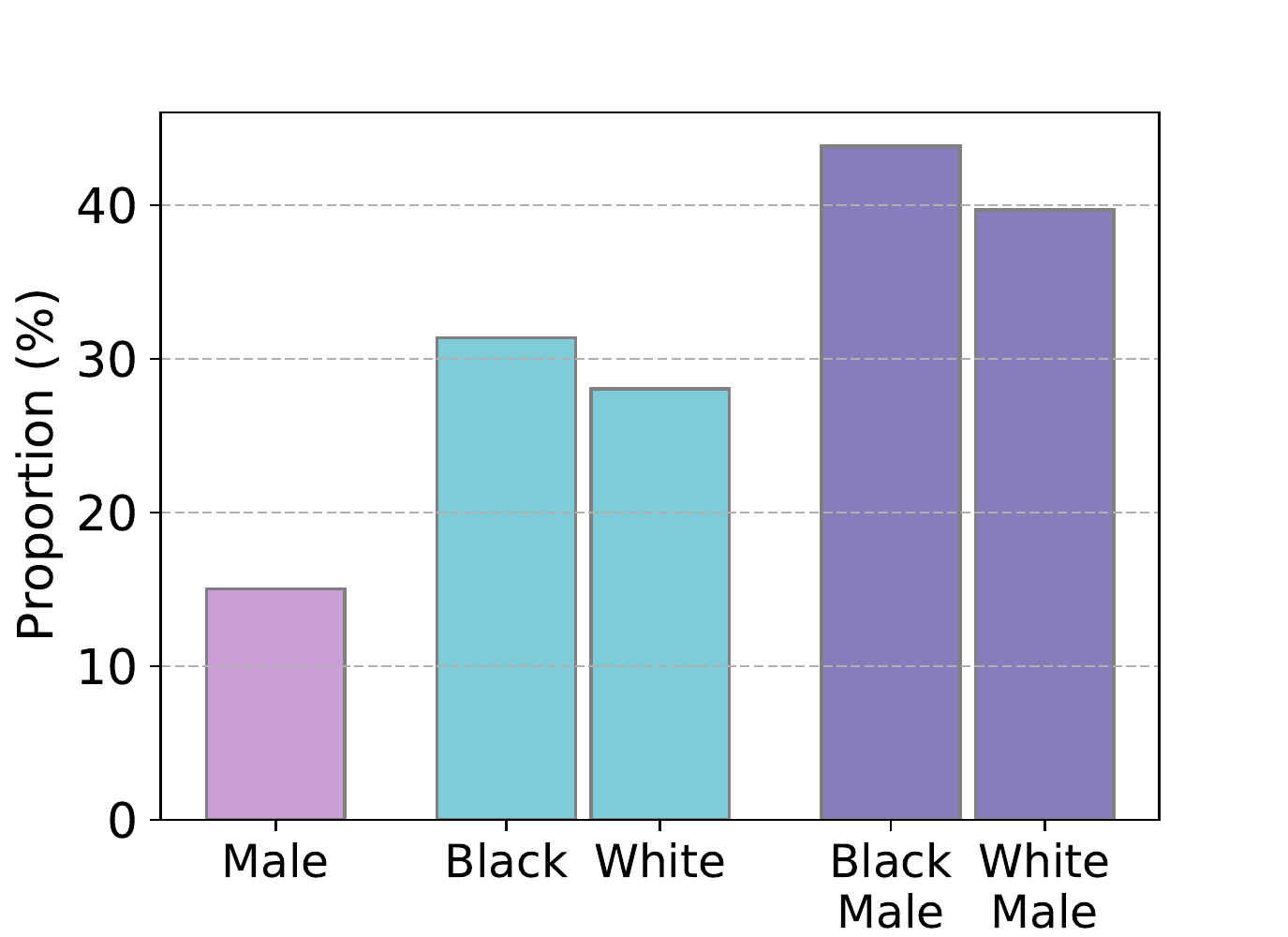}}\label{subfig:prop}} \qquad
\subfigure[FPR of the \textit{Joint} model.]{\resizebox{0.28\textwidth}{!}{\includegraphics{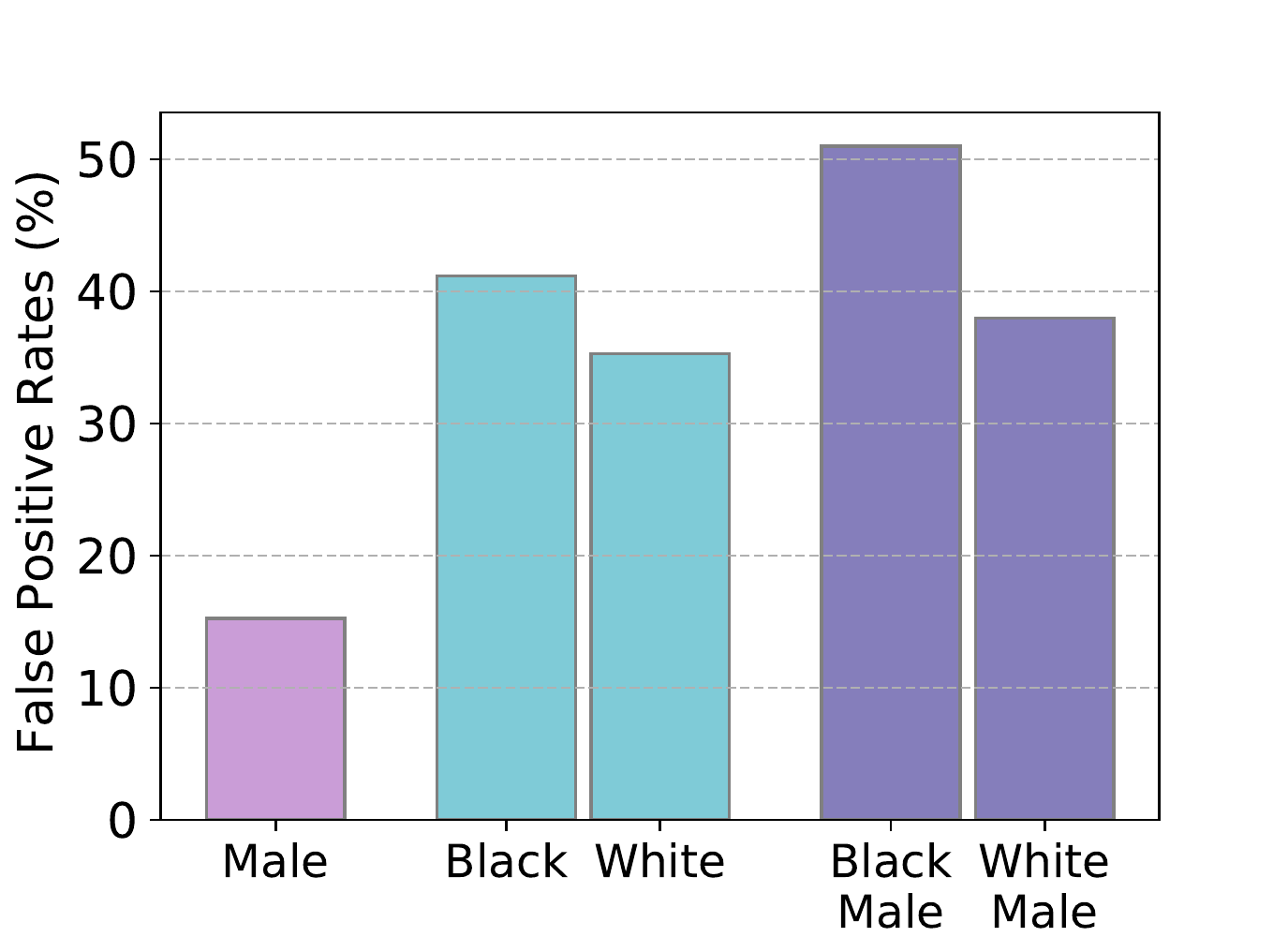}}\label{subfig:fpr}} \qquad
\subfigure[FNR of the \textit{Joint} model.]{\resizebox{0.28\textwidth}{!}{\includegraphics{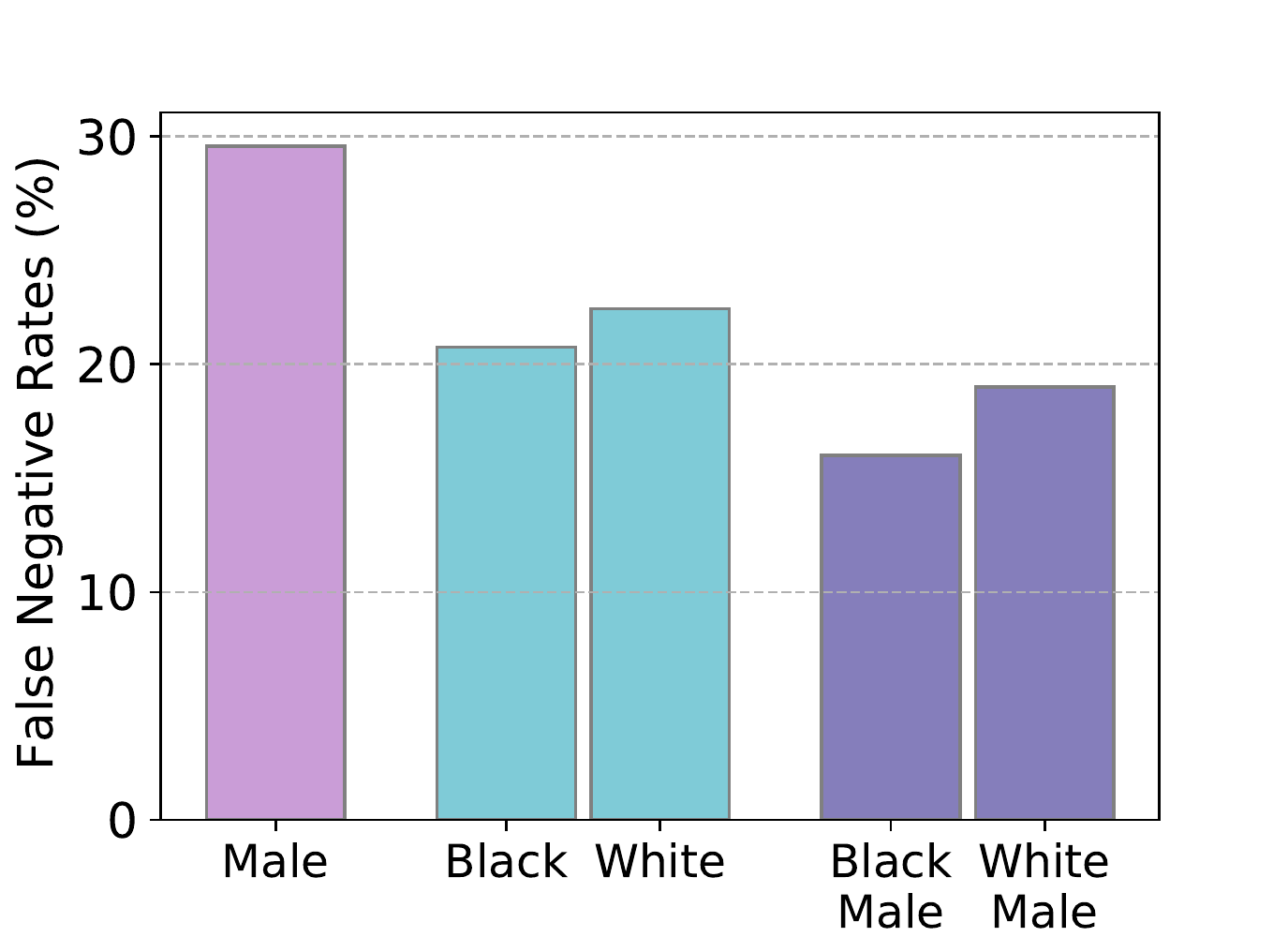}}\label{subfig:fnr}}
\caption{Potential explanations to bias correlations in debiasing toxicity detection.
Fig.~\ref{subfig:prop} shows the imbalanced distribution of toxicity annotations in the original \textit{Jigsaw} datasets.
We observe that the toxicity proportions of samples annotated with intersectional identities are higher.
Fig.~\ref{subfig:fpr} and \ref{subfig:fnr} elucidate that \textit{Joint} tends to further amplify such imbalance by more likely predicting samples with intersectional identities as ``toxic''.
}
\label{fig:proportionfprfnr}
\end{figure*}
\noindent\textbf{RQ. 2.} To see the effectiveness of \textit{Joint}, we compare joint debiasing with sequential debiasing. In particular, we repeat the experiments in Table \ref{rq1}-\ref{rq2} and obtain the results for \textit{Joint}. We observe that in Table \ref{rq1}, \textit{Joint} outperforms sequential debiasings and often achieves better performance than methods that debiasing exclusively for the target identities. Take the last row as an example where gender is the target identity, \textit{Joint} (.790) removes greater gender biases than \textit{H-Ge} (.695) and \textit{H-Re-Ge} (.654). Similarly, on row 3 in Table \ref{rq2}, jointly debiasing for religion and gender (.790) and all three identities (.794) remove larger biases than the corresponding sequential debiasing methods, i.e., \textit{H-Re-Ge} (.654) and \textit{H-Re-Ra-Ge} (.655). In addition to the quantitative measures, we further generate the top five analogies for \{\textit{man, woman}\} using various word embeddings debiased for gender. We observe from Table \ref{analogy} that \textit{H-Ge} and \textit{Joint} generate same analogies for both \textit{man} and \textit{woman}. The sequential debiasing methods (i.e., \textit{H-Re-Ge} and \textit{H-Re-Ra-Ge}), however, generate discriminative analogies as highlighted.

\noindent\textbf{RQ. 3.} We follow \cite{manzini2019black} to examine the effects of mitigating biases on three downstream tasks: NER Tagging, POS Tagging, and POS Chunking. Specifically, we compare the utility of \textit{Biased} with word embeddings debiased at three different levels: individual identity (\textbf{Re}ligion), two (\textbf{Re}ligion and \textbf{Ra}ce), and all three identities. We report results of embedding matrix replacement and model retraining in Table \ref{downstream1}-\ref{downstream2}. We observe that for all word embeddings, the semantic utility only slightly changes. Student $t$ test further testifies that these differences are insignificant. We conclude that hard-debiasing does not have significant influence on the utility of word embeddings. This applies to both sequential and joint bias mitigation.  
\subsection{Potential Explanations to Bias Correlations}
In this section, we take a closer look at the potential explanations to the observed positive/negative bias correlations. Particularly, we show both quantitative results and documented findings related to bias correlations in social psychology. 
\begin{figure}[t]
    \centering
\subfigure[Bias correlation in 3D.]{\resizebox{0.5\columnwidth}{!}{\includegraphics{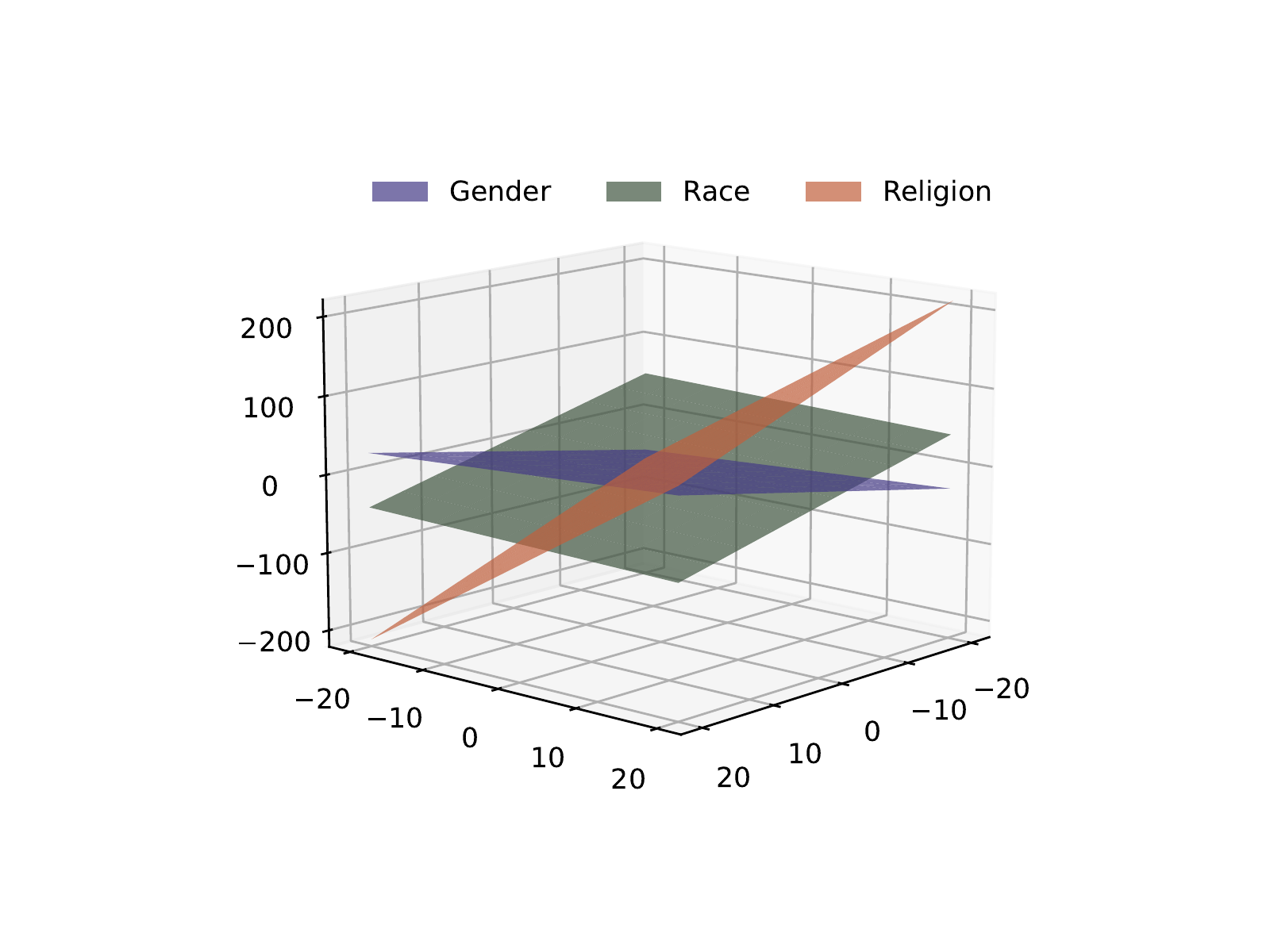}}}
\subfigure[Bias correlation in 2D.]{\resizebox{0.45\columnwidth}{!}{\includegraphics{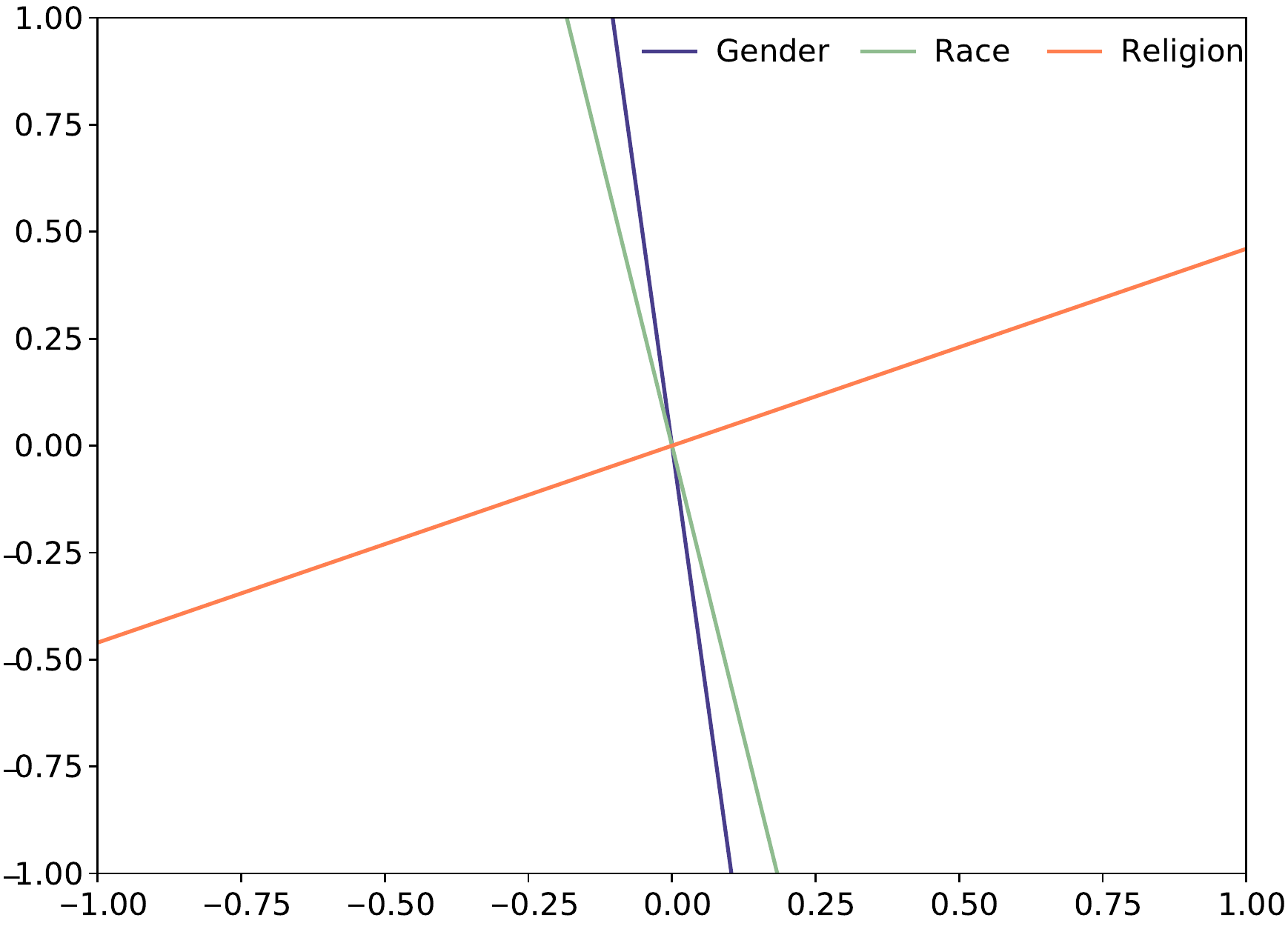}}}
\caption{Visualization of the high-dimensional Gender, Race, and Religion bias subspaces.}
\label{subspaces_overlap}
\end{figure}
\subsubsection{Toxicity Detection.}
We first examine the distribution of toxicity labels in different social identities in Fig.~\ref{subfig:prop}.
We present the toxicity proportions of groups with single identity (male, black, and white) and groups with intersectional identities (black male and white male).
First, the higher toxicity proportions of black male and white male in the annotated data indicate that intersectional identities may potentially contribute to larger prominent social bias than the constituent identities.
We then investigate whether similar findings can be found at the model level. Fig.~\ref{subfig:fpr} and \ref{subfig:fnr} show the FPR and FNR calculated based on prediction results of \textbf{Joint}.
We observe that groups with intersectional identities show higher FPR (falsely identified as toxic) and lower FNR (falsely identified as non-toxic) than the constituent individual identities. Such imbalance tends to be more significant at the model level. This indicates that there is a larger bias against intersectional identities than the constituent identities and bias correlations observed in the model inherit from or even amplify bias correlations in the data.
\subsubsection{Word Embeddings.}
To get a sense of why biases might be correlated with each other in debiasing word embeddings, we visualize the three identified bias spaces (i.e., gender, race, and religion). In particular, we first reduce each bias subspace to 3D vectors using PCA and then visualize these subspaces in low-dimension spaces in Fig. \ref{subspaces_overlap}(a)-(b), respectively. We observe that all the three subspaces intersect with each other and the angles of these intersections are different. For instance, $\measuredangle$ (Gender, Race) $<$ $\measuredangle$ (Gender, Religion), that is, gender and race tend to be more correlated. The intersection might explain the interdependent neutralizing operations in the sequential debiasings. As for the ``negative'' correlation, one explanation is because these subspaces are positioned in opposite directions, bias removals can be cancelled out when adjusting for the bias subspaces sequentially. 
\subsubsection{Findings in Social Psychology} The interrelationship between various biases have been one of the oldest lessons in the prejudice literature, which was referred to as ethnocentrism \cite{adorno1950authoritarian}. A more common term used in the modern social psychology is ``generalized prejudice'' \cite{allport1954nature}, defined as ``generalization of devaluing sentiments across different group domains, as reflected in broad individual difference coherences.'' \cite{bergh2016group} It suggests that people who discriminate one group domain tend to also discriminate other group domains. For example, if a person devalues ethnic minorities, s/he is likely to devalue women, gays and a whole range of other groups \cite{bergh2016group}. Findings indicating such positive correlations have been documented numerous times across a range of cultural contexts (e.g., \cite{akrami2011generalized,bierly1985prejudice,backstrom2007structural}). Indeed, research has found prejudice toward various group domains to be significantly correlated, e.g., a generalized prejudice factor explaining 50\% to 60\% of the variance as shown in a factor analysis \cite{ekehammar2003relation}. Therefore, bias can be generalized across groups. Bias correlations in AI may further stem from other factors. For example, AI researchers and practitioners who do not take active measures to try and mitigate one form of bias tend to not take measures to reduce other forms of bias. Although two distinct forms of bias are inherently correlated as shown in social psychology findings, we further provide empirical evidence of negative bias correlations in debiasing. This research opens up potential intersections between NLP and quantitative social psychology.

\subsection{Discussions}
Based on the empirical evaluations and prior research in social psychology, we summarize the \textbf{key findings} toward understanding bias correlations in bias mitigation: (1) Social biases are correlated and correlations can be positive or negative, depending on the debiasing levels (e.g., data or model level) and the domain applications. For example, certain identity-based dialects may be more likely to be flagged as having toxic content (due to identity-related text) and the dialect associated with certain identity groups may be more likely to contain swear words \cite{zhou2021challenges}. Under positive correlations, prior works that treat biases independently may still show satisfactory performance w.r.t. mitigating the total bias. However, our results suggest that biases can also be negatively correlated during the debiasing process. Therefore, it is of considerable practical importance to account for the correlations in bias mitigation. (2) Regardless of the correlation being positive or negative, a joint bias mitigation approach is more effective in reducing the total bias than debiasing independently. Additionally, the improvement appears to be more significant under negative bias correlations. Therefore, we advocate the need to jointly debias for various social identities, especially when the bias correlation is negative. (3) Although debiasing may not have influence on the utility (e.g., debiaisng word embeddings) or even improve the prediction performance (e.g., debiasing toxicity detection), there might exist an inherent trade-off between debiasing and utility during model training.

The study is not without limitations. 
First, there might be other sources contributing to data biases, such as the selection bias during data collection and annotation bias due to the diverse background of annotators.
Second, while interesting and important, our findings are made based on the results for limited datasets and tasks. Future research can be directed towards investigating other NLP models/tasks (such as contexutlized word embeddings, pre-trained language models, token-level or text generation tasks) using different baselines and datasets to examine the generalizability of our findings.
Third, given the potential negative results of independent debiasing, future efforts need to be focused on developing more principled approaches and evaluation metrics for joint bias mitigation. For example, by customizing different bias correlations, a joint debiasing method may achieve better mitigation performance and alleviate the issue of debiasing-accuracy trade-off.
\section{Conclusions}
This work provides an in-depth understanding of bias correlations for bias mitigation in NLP. It serves to bring forefront the discussions of the potential limitations in current independent debiasing approaches that might be overlooked before. By investigating two common NLP tasks (i.e., mitigating bias in word embeddings and toxicity detection), we show that biases are correlated, echoing the ``generalized prejudice'' theory in social psychology. This study also complements this theory suggesting positive correlations by showing that social biases can also be negatively correlated. Therefore, we further advocate to develop joint bias mitigation approaches, which we showed that outperform independent debiasing under both scenarios of positive and negative bias correlations. Lastly, in spite of the promising results of \textit{Joint}, it also presents inherent debiasing-accuracy trade-off during the model training process. 
\bibliographystyle{ACM-Reference-Format}
\bibliography{sample-base}
\clearpage
\appendix
\section{Appendices}
\subsection{Parameter Settings}
\subsubsection{Debiasing Toxicity Detection}
For debiasing toxicity detection with constraints, we use the publicly available implementation in ~\cite{gencoglu2020cyberbullying}\footnote{https://github.com/ogencoglu/fair\_cyberbullying\_detection}.
In particular, we employ a multilingual language model, sentence-DistilBERT~\cite{reimers2019sentence}, for extracting sentence embeddings to represent each post/comment. The toxicity classifier is a simple 3-layer fully-connected neural network. 
The size of each layer is 512, 32, and 1, respectively.
We use sentence embeddings output from sentence-DistilBERT as input features.
The model is trained for 25 epochs with Adam as the optimizer.
In the repository, they also release their trained models along with the source code.
To account for the performance variance caused by the system environment, we ran the source code from scratch and reported our experimental results.
Both baselines and the constrained models are trained in a mini-batch manner. 
Models that maximize the F1 score on the validation set are used for experimentation. 
We ran each experiment for five times and reported the average performances.
We describe the details of major parameter setting in Table \ref{tab:toxicity}. The descriptions of the major parameters are as follows:
\begin{itemize}[leftmargin=*]
    \item \textit{Embedding Dimension}: the dimension of BERT sentence embeddings.
    \item \textit{LR Constraints}: the updating rate of proxy-Lagrangian state.
    \item \textit{Adam\_beta\_1}: the exponential decay rate for the 1st moment estimates.
    \item \textit{Adam\_beta\_2}: the exponential decay rate for the 2nd moment estimates.
    \item \textit{FNR Deviation}: the maximum allowed deviation for FNR.
    \item \textit{FPR Deviation}: the maximum allowed deviation for FPR.
\end{itemize}
\subsubsection{Debiasing Word Embeddings}
For hard debiasing word embeddings, the major parameter $k$ (the most $k$ significant components in PCA) is set to 2 for all bias subspace identifications. 
We use the same data split strategy as~\cite{manzini2019black} by randomly splitting the dataset into 80\% for training, 10\% for validation, and 10\% for testing. 
We also follow their parameter settings for all downstream tasks. 
We ran each experiment for five times and reported the averaged performances.
The major parameters in debiasing word embeddings are detailed as follows:
\begin{itemize}[leftmargin=*]
    \item \textit{Max Seq Len}: the threshold to control the maximum length of sentences.
    \item \textit{Embedding Size}: the dimension of embedding layer.
    \item \textit{Debias\_eps}: the threshold for detecting words that had their biases altered.
\end{itemize}
We summarize major parameter settings in Table \ref{tab:word}.

\subsection{Ethics Statement}
This work aims to advance collaborative research efforts in understanding bias correlations and joint bias mitigation in NLP. Here, we provide the first systematic study to examine the relations between biases and further propose two simple solutions to mitigating multiple correlated biases. However, much work remains to be done to understand ``why'' biases are correlated and to build a more effective framework in the presence of correlated biases. All data in this study are publicly available and used under ethical considerations. Text and figures that contain terms considered profane, vulgar, or offensive are used for illustration only, they do not represent the ethical attitude of the authors.
\begin{table}[t]
	\caption{Details of the parameters in the debiasing toxicity detection experiment.}
	\centering
	\resizebox{0.47\textwidth}{!}{
	\begin{tabular}{l|l||l|l}
    \Xhline{1pt}
            Parameter&Setting&Parameter&Setting\\
            \hline
			Batch Size&128&Embedding Dimension&512\\
			\hline
			Learning Rate (LR)&5e-4&LR Constrains&5e-3\\
			\hline
			Adam\_beta\_1 &0.9& Adam\_beta\_2 &0.999\\
			\hline
			FNR Deviation&0.02&FPR Deviation&0.03\\
    \Xhline{1pt}
	\end{tabular}}

	\label{tab:toxicity}
\end{table}

\begin{table}[t]
	\caption{Details of the parameters in the debiasing word embedding experiment.}
	\centering
	\resizebox{0.47\textwidth}{!}{
	\begin{tabular}{l|l||l|l}
    \Xhline{1pt}
            Parameter&Setting&Parameter&Setting\\
    \hline
            Max Seq Len&128&Embedding Size&50\\\hline
			Learning Rate (LR)&1e-3&Epochs&25\\\hline
			RMSprop Decay&1e-3& RMSprop Momentum&0.25\\\hline
			Debias\_eps & 1e-10& Batch Size&64\\
    \Xhline{1pt}
	\end{tabular}}

	\label{tab:word}
\end{table}


\end{document}